\documentclass[runningheads]{llncs}
\usepackage{makeidx}
\usepackage{graphicx}
\usepackage{amsmath,amssymb} 
\usepackage{color}
\usepackage{xspace}
\usepackage{gensymb}
\usepackage[svgpath=./images]{svg}
\usepackage{hyperref}
\usepackage{pdfpages}

\newcommand\blfootnote[1]{%
  \begingroup
  \begin{tiny}
  \renewcommand\thefootnote{}\footnote{#1}%
  \addtocounter{footnote}{-1}%
  \end{tiniy}
  \endgroup
}

\newcommand{\dataMixed}{$\mathcal{D}_M$\xspace}
\newcommand{\dataStyle}{$\mathcal{D}_{Style}$\xspace}
\newcommand{\dataReal}{$\mathcal{D}_{R}$\xspace}
\newcommand{\dataSynth}{$\mathcal{D}_{S}$\xspace}
\newcommand{\camPosGroup}{$C$\xspace}
\newcommand{\minDistGroup}{$mD$\xspace}
\newcommand{\supmat}{{\bf{Supp. Mat.}}\xspace}
\newcommand{\dataSynthAndReal}{$\mathcal{D}_{R}$+$\mathcal{D}_S$\xspace}
\newcommand{\dataMixedAndReal}{$\mathcal{D}_{R}$+$\mathcal{D}_M$\xspace}
\newcommand{\dataStylizedAndReal}{$\mathcal{D}_{R}$+$\mathcal{D}_{Style}$\xspace}

\newcommand{\baseline}{$\mathcal{M}_{\mathcal{D}_{R}}$\xspace}
\newcommand{\realAndSynth}{$\mathcal{M}_{\mathcal{D}_{R}+\mathcal{D}_{S}}$\xspace}
\newcommand{\synthOnly}{$\mathcal{M}_{\mathcal{D}_{S}}$\xspace}
\newcommand{\mixed}{$\mathcal{M}_{\mathcal{D}_{R}+\mathcal{D}_M}$\xspace}
\newcommand{\stylized}{$\mathcal{M}_{\mathcal{D}_{R}+\mathcal{D}_{Style}}$\xspace}
\newcommand{\mixedMasked}{$\mathcal{M}_{\mathcal{D}_{R}+\mathcal{D}_M+\text{masks}}$\xspace}
\newcommand{\stylizedMasked}{$\mathcal{M}_{\mathcal{D}_{R}+\mathcal{D}_{Style}+\text{masks}}$\xspace}

\newcommand{\advTeacherTd}{``adversarial Teacher''\xspace}

\newcommand{\new}[1]{\textcolor{black}{#1}}

\begin{document}
	\pagestyle{headings}
	\mainmatter
 
	\def\GCPR19SubNumber{132}

	\title{Learning to Train with Synthetic Humans}
	
	\titlerunning{Learning to Train with Synthetic Humans}
	\authorrunning{D.~T.~Hoffmann, D.~Tzionas, M.~J.~Black,  S.~Tang}
	\author{    David T.~Hoffmann      \inst{1} 
	            \and Dimitrios Tzionas  \inst{1}   
	            \and Michael J.~Black   \inst{1}   
	            \and Siyu Tang          \inst{1,2,3} 
	}
	\institute{ $^1$ Max Planck Institute for Intelligent Systems, Germany         \\
	            $^2$ University of T{\"u}bingen, Germany                           
	            $^3$ ETH Z\"urich\\
	            \email{{dhoffmann, dtzionas, black, stang}@tuebingen.mpg.de}
    }
	\maketitle

	\begin{abstract}
        Neural networks need big annotated datasets for training. However, manual annotation can be too expensive or even unfeasible for certain tasks, like multi-person 2D pose estimation with severe occlusions.
        A remedy for this is synthetic data with perfect ground truth. 
        Here we explore two variations of synthetic data for this challenging problem;  
        a dataset with purely synthetic humans and a real dataset augmented with synthetic humans.
        We then study which approach better generalizes to real data, as well as the influence of virtual humans in the training loss. 
        \new{Using the augmented dataset, without considering synthetic humans in the loss, leads to the best results.}
        We observe that not all synthetic samples are equally informative for training, while the informative samples are different for each training stage. 
        To exploit this observation, we employ an adversarial student-teacher framework; 
        the teacher improves the student by 
        providing the hardest samples for its current state as a challenge. 
        \new{Experiments show that the student-teacher framework outperforms normal training on the purely synthetic dataset.}
	\end{abstract}

\section{Introduction}
\label{sec:introduction}

\begin{figure}[t]
    \centering

    \caption{\new{
    Qualitative comparisons between our models and Cao et al.~\cite{cao2017}.
    }}
    \includegraphics[width=\textwidth]{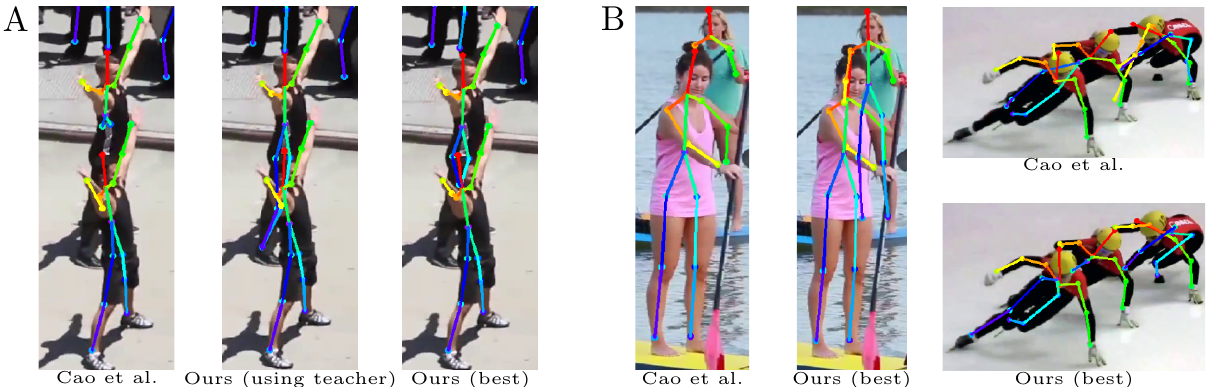}
    \label{fig:teaser}
\end{figure}

The broad success of deep neural networks comes at a price: the ever growing need for huge amounts of labeled training data. 
For many tasks, the lack of data seems to be one of the major limiting factors of progress. 
It is particularly problematic for the tasks where manual labeling requires significant human effort, or is even unfeasible. For example, in multi-person 2D pose estimation, a major challenge is that people are often partially visible. Manual annotation of body joints that are severely occluded is error prone and the resulting labels are noisy. 
Computer graphics can help to resolve these issues.
3D rendering engines offer the opportunity to generate a \new{large} amount of
data with perfect labels:
e.g., the location of occluded body parts and the precise pose of the camera.

Nowadays, large scale synthetic datasets with reasonable realism can be generated relatively easy and the idea of synthesizing training data has been widely explored. In general, there are two common strategies for generating a synthetic dataset: rendering a purely synthetic dataset, and augmenting real training images with synthetic instances.
The advantage of the former is the full control over the virtual 3D world and ability to generate high variance datasets \cite{bak2018domain,barbosa2018looking,fabbri2018learning,marin2010learning,pishchulin2011learning,Ranjan_BMVC_2018,rogez2019lcr,varol17_surreal}.
The advantage of the second approach is that some of the instances in the dataset are real, resulting in overall higher realism \cite{alhaija2018augmented,dvornik2018importance,sarandi2018robust}.

We generate both types of synthetic datasets. One purely synthetic dataset and a mixed dataset, which is generated by augmenting the MPII pose estimation dataset \cite{andriluka14cvpr} with synthetic humans. 
In particular, we design these datasets to improve on frequent failure cases that we observe with state-of-the-art models (see Fig~\ref{fig:teaser}), namely uncommon camera angles and strong occlusion.
By comparing generalization performance using these two datasets we obtain insights into which way of generating data is preferable.
We further investigate how strongly the lack of photorealism of the synthetic humans limits  generalization. To make the synthetic data more realistic, we propose a simple synthetic-to-real human style transfer algorithm, based on the work of Dundar et al.~\cite{dundar2018domain}.

These experiments show that naive training with synthetic data leads only to limited improvements.
One explanation is that training on large synthetic datasets leads to overfitting of the model to the features of synthetic data.
We observe that some synthetic images convey more information than others.
Overfitting to features of synthetic data could be limited by generating only useful, i.e.~difficult synthetic data, and thus limiting the training on synthetic data.

As a step in this direction,
we propose a method to use synthetic datasets more effectively. 
Specifically, we introduce an adversarial student-teacher framework.
The teacher learns online which training data is still difficult.
This information is then used to increase the sampling probability of similar examples.
By taking into account feedback from the student, the teacher keeps on updating the sampling probabilities throughout training and adapts them to the specific needs of the student.
\new{Training with the teacher on the purely synthetic data outperforms normal training.}

Our contributions can be summarized as follows:
1) We propose a large-scale synthetic multi-person dataset, a mixed dataset, and a domain-adapted version of the latter. 2) We explore which way of generating synthetic data is superior for our task. 
3) We propose a student-teacher framework to train on the most difficult images and show that this method outperforms random sampling of training data on the synthetic dataset.
\new{We provide datasets and code\footnote{\url{https://ltsh.is.tue.mpg.de}}}.
\section{Related Work}

{\bf Synthetic datasets with humans.} The need for labelled training data has fueled development of datasets with synthetic humans.
Many methods use 3D models of the human body to generate data \cite{bak2018domain,barbosa2018looking,ghezelghieh2016learning,pishchulin2011learning}. 
Other approaches augment 3D training data by utilizing 2D pose datasets \cite{chen2016synthesizing,rogez2018image},
while \cite{sarandi2018robust,tripathi2019learning} augment datasets with cut-outs of objects or animals.
Closely related to our approach, \cite{Ranjan_BMVC_2018,varol17_surreal}  
render the SMPL model \cite{loper2015} on top of random indoor images.
\new{These methods} generate datasets with a single synthetic human.
\new{Multi-person datasets were created by employing video games for pedestrian detection \cite{marin2010learning} and pose tracking \cite{fabbri2018learning}.}
Similarly, \cite{muller2018sim4cv} develop a simulation environment in a game engine, including virtual humans.
Related to our approach,
\cite{rogez2019lcr} augment real training images similar to the ones in 
\cite{varol17_surreal} but with multiple synthetic humans occluding each other. 

{\bf Domain Adaptation.} The quality of synthetic data is often insufficient to generalize well to real data. 
Several domain adaptation methods have been developed to overcome this problem.
Shrivastava et al.~\cite{shrivastava2017learning} 
train a Generative Adversarial Network (GAN) to refine synthetic images, while keeping the label information intact.
Recently, Cycle-GAN has been used to map images from one domain to another \cite{bak2018domain,murez2018image}.
However, GAN-based methods are prone to unstable training and require tedious hyper-parameter tuning.
As a practical alternative, recently 
\cite{dundar2018domain} proposed domain stylization to stylize synthetic images to real ones, using the fast photorealistic image stylization method of \cite{li2018closed}. 

{\bf Human pose estimation.} Multi-person pose estimation has attracted substantial attention over the last years \cite{cao2017,fang2017,fieraru2018learning,insafutdinov2016deepercut,kocabas2018multiposenet,luo2019multi,newell2017,nie2018pose}. 
One of the most popular 
datasets is the MPII multi-person pose estimation dataset \cite{andriluka14cvpr}. 
Among the best performing methods on MPII are: \cite{nie2018pose}, which uses a pose partition network, 
\cite{luo2019multi}, which uses context information,
\cite{fieraru2018learning}, which refines pose predictions, and
\cite{newell2017}, which predicts ``tag maps'' to solve the grouping problem. 
The most widely used 
method is OpenPose 
\cite{cao2017}, a bottom-up approach that first predicts keypoints and then estimates Part Affinity Fields (PAFs) to group them.

{\bf Learning to train.} Bengio et al.~\cite{bengio2009curriculum} introduce curriculum learning for learning systems, exploiting the idea that different data samples are informative at different training stages. As a proof of concept they manually define the samples for each stage with gradually increasing difficulty. 
Multiple methods have focused on automating curriculum learning  
\cite{buchler2018improving,fan2017learning,jiang2017mentornet,kumar2010self}. 
These approaches try to maximize information gain during training by carefully monitoring the learning success of the model.
Alternatively, adversarial methods  \cite{katharopoulos2017biased,kim2018screenernet,shrivastava2016training}
try to pick the hardest samples at each training stage; 
\cite{katharopoulos2017biased} favors samples resulting in higher loss, 
\cite{kim2018screenernet} learns weights for the loss of each training sample as a soft curriculum, while 
\cite{shrivastava2016training} uses online hard example mining for object detection.
Peng et al.~\cite{peng2018jointly} propose an adversarial training scheme to optimize data augmentation online.
They train two teacher networks to learn a probability distribution over the hyper-parameters for data augmentation;
one predicts the most difficult image rotations, while the other predicts parameters for deep feature occlusion.
\section{Multi-person Synthetic Data}

In the following we describe the generation of the datasets (Sections \ref{sec:dataGenerationPipeline}, \ref{sec:syntheticDataset}, \ref{sec:mixedDataset}) and the domain adaptation used to increase visual appearance of the mixed dataset (Section \ref{sec:domainStylization}).

\subsection{Data Generation Pipeline}   \label{sec:dataGenerationPipeline}
\begin{figure}[t]
    \centering
    \caption{Schematic of the data generation pipeline.}
	\includegraphics[width=0.9\textwidth]{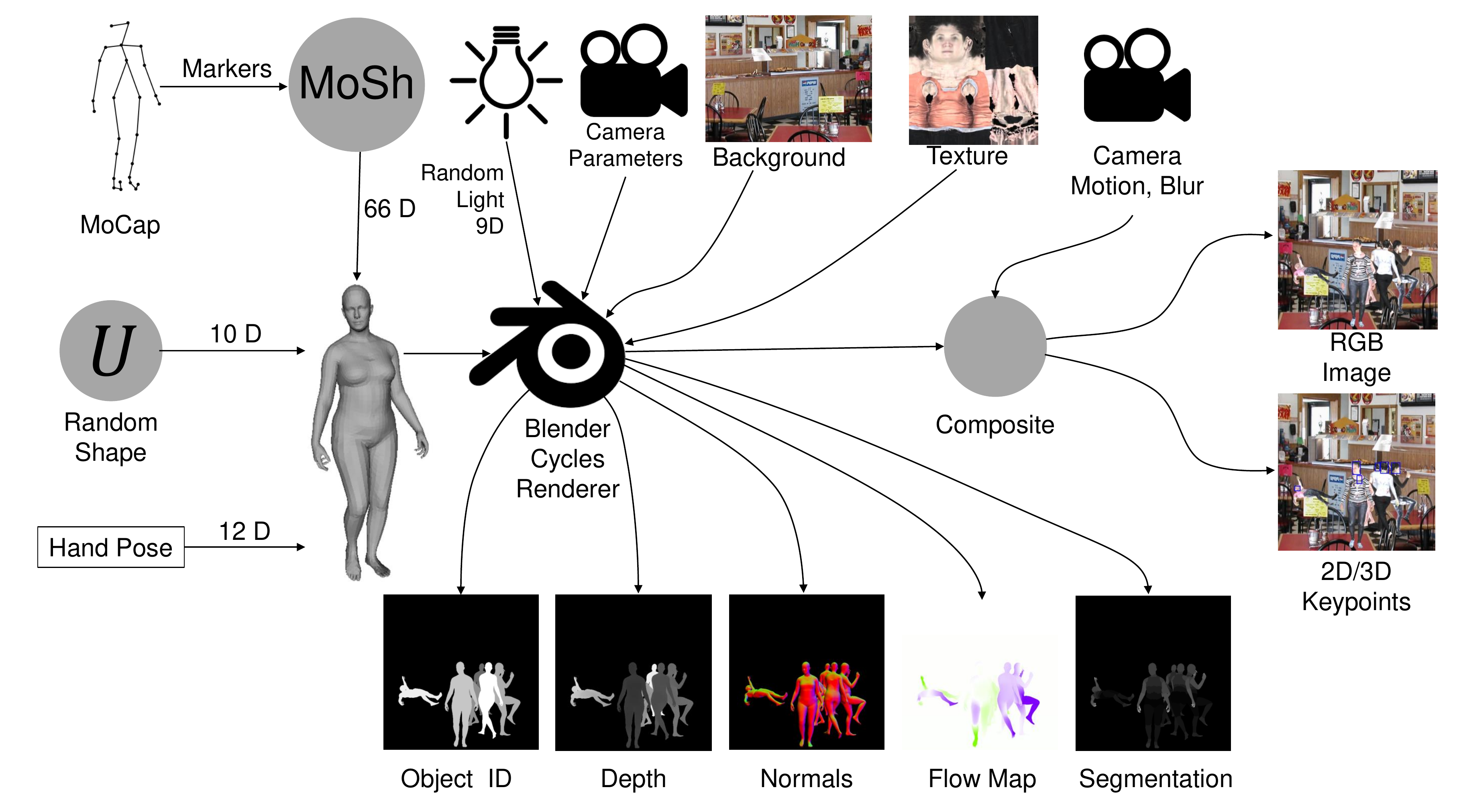}
    \label{fig:data_gen_pipe}
\end{figure}

To generate realistic synthetic training data we build on top of \cite{Ranjan_BMVC_2018,varol17_surreal}.
We use multiple different sources of data to build a realistic synthetic scene. Images and ground truth annotations are rendered using Cycles, the rendering engine of Blender\footnote{\url{https://www.blender.org}}.
An overview can be seen in Fig. \ref{fig:data_gen_pipe}. 

{\bf Body Model and MoCap Data.} We use the parametric body model SMPL+H \cite{MANO:SIGGRAPHASIA:2017} to generate realistic synthetic humans. 
SMPL+H is parameterized by pose $\theta \in \mathbb{R}^{78}$ and shape $\beta \in \mathbb{R}^{10}$. 
We collect realistic pose and shape parameters by fitting SMPL+H to standard MoCap data by using MoSh \cite{loper2014mosh}.

{\bf Details.} We draw inspiration from \cite{Ranjan_BMVC_2018,varol17_surreal} who generate small video sequences, each having different but fixed parameters for the position, pose, shape and texture of a synthetic human, the background image, camera position, lighting, etc. 
In contrast, we generate single images 
and render multiple synthetic humans, while randomizing the number of them.
As a result, we generate a dataset with much higher variance.
Images with inter-penetrating meshes of virtual humans are rejected to avoid artifacts in the generated ground truth. 

Further details regarding the description of the data generation pipeline, posing of hands and a quantitative comparison of our datasets to other datasets can be found in \supmat

\begin{figure}[t]
    \centering
    \caption{Example images from the purely synthetic dataset. It contains high occlusion, extreme poses, various camera angles and various challenging backgrounds.}
    \includegraphics[width=0.9\textwidth]{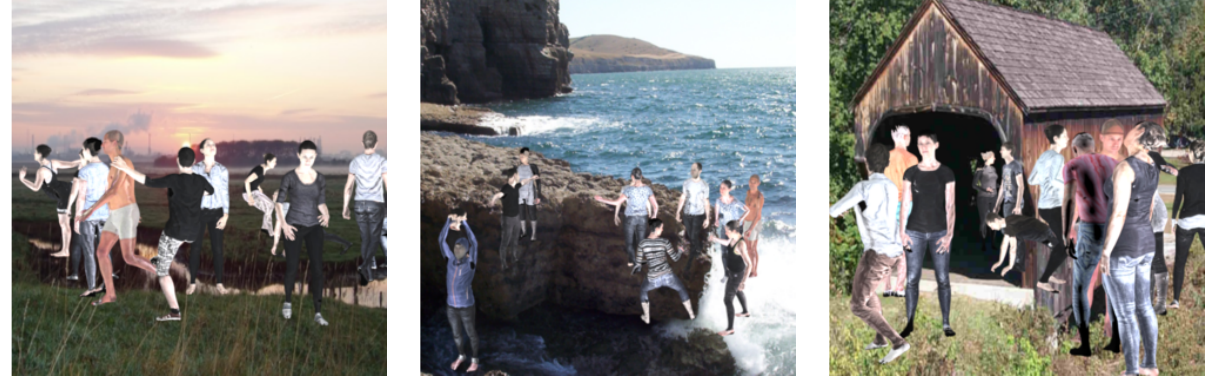}
    \label{fig:my_label}
\end{figure}

\subsection{Synthetic Dataset}  \label{sec:syntheticDataset}

{\bf Background Images.} To generalize well to in-the-wild pose estimation datasets the background images should come from many different scenes.
To this end we use images from SUN397 \cite{xiao2010sun} and reject all images with a resolution smaller than $512 \times 512$ pixels to ensure high quality backgrounds. Additionally, we reject all images containing humans, as we do not have ground truth annotations for them. We use mask-RCNN \cite{matterport_maskrcnn_2017,He2017} as our human detector.

{\bf Generative Factors.} We sample the number of synthetic humans per image from a Poisson distribution with $\lambda=9$, to encourage many humans per image, while avoiding too extreme values.
The datasets of \cite{Ranjan_BMVC_2018,varol17_surreal} have only very small variance in camera position.
However, preliminary experiments show that the camera position significantly influences the difficulty of multi-person pose estimation. 
Therefore we increase the range of possible camera positions, by sampling the camera pitch uniformly from $[0,45]^\circ$.
The resolution of the final rendered images is set to $640 \times 640$ pixels.
We refer to this dataset as \dataSynth.

\subsection{Mixed Dataset}  \label{sec:mixedDataset}
\begin{figure}[t]
    \centering
    \caption{(A) Example images from \dataMixed. (B) Corresponding images of \dataStyle. For the last image, the segmentation network included non-human parts in the segmentation masks. Resulting artifacts can be seen for rightmost synthetic human.}
    \includegraphics[width=0.9\textwidth]{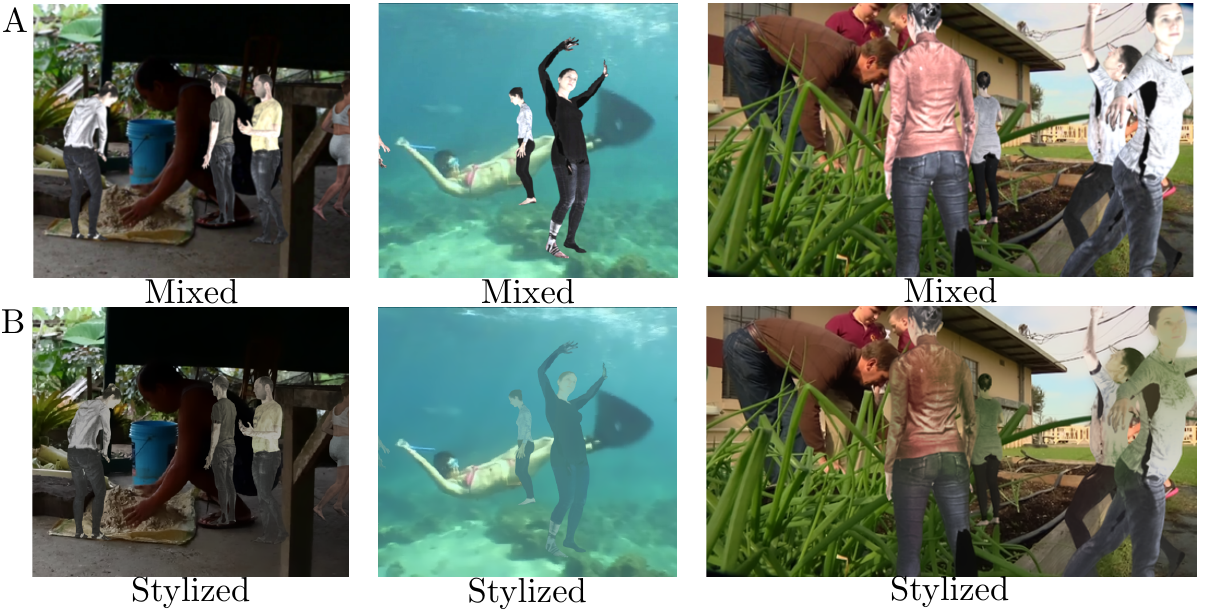}
    \label{fig:mixed_style}
\end{figure}

\new{
We build upon the finding of \cite{sarandi2018robust} that realistic occluding objects lead to larger improvements than abstract objects. 
We choose our occluders to be from the same class as our target objects; i.e.~humans, to simulate crowded scenes with multiple humans.}
To generate the dataset we use the pipeline described above with a few differences.
Instead of SUN397 \cite{xiao2010sun} we use the training images of the MPII human pose dataset \cite{andriluka14cvpr} as background images.
To keep the MPII ground truth intact, we render the images with the same resolution as the background MPII image,
and keep the camera pose fixed. 
We then augment the MPII human pose dataset by superimposing synthetic humans. Their number is drawn from a Poisson distribution with $\lambda=4$ 
to introduce interesting and intense occlusions 
as shown in Fig.~\ref{fig:mixed_style} (A), 
without extreme occlusions by too many synthetic humans.
We render each of the  $15,956$ images in our training set $5$ times with different parameters for increased variance. 
We refer to this dataset as \dataMixed.

\subsection{Domain Stylization}     \label{sec:domainStylization}
The appearance of real and synthetic humans differs strongly. Factors contributing to these differences are the low quality of textures and differences in lighting conditions for synthetic humans and background images. 
Additionally, the small number of human textures limits the variability.
These differences in appearance might limit the generalization. 
We draw inspiration from Dundar et al. \cite{dundar2018domain} to reduce these differences by using the fast photorealistic style transfer method of \cite{li2018closed}.
Style transfer methods require a pair of images as input, a content image $I_C$ and a style image $I_S$. While the style of these images can be largely different, their content should have similarities. 
Finding such pairs is a non-trivial problem.
However, for the \dataMixed, we have a canonical choice of image pairs: the image from \dataMixed and its background as $I_C$ and $I_S$, respectively.

Naive application of style transfer methods on the whole image, leads to severe artifacts. 
Therefore, only the style of semantically similar classes should be transferred. 
To obtain a good semantic segmentation network Dundar et al.~\cite{dundar2018domain} iterate between stylizing a dataset and training a network for semantic segmentation on the stylized dataset.
Here, we are only interested in transferring the style of real to synthetic humans.
Fine-grained human detection is important to avoid parts of the background bleeding into the foreground after style transfer (see Fig.~\ref{fig:mixed_style} (B) right panel). 
We therefore employ Mask-RCNN \cite{matterport_maskrcnn_2017,He2017} to predict pixel-wise masks for humans. 
Ground truth masks for the synthetic humans are generated during data generation. 
Since the style-transfer algorithm can not handle images of arbitrary size, we rescale the larger images to $600$ pixels before applying the style transfer.
We refer to the resulting dataset as \dataStyle.
Examples can be seen in Fig.~\ref{fig:mixed_style} (B).

\section{Learning to Train with Multi-person Synthetic Data}

We now turn to the task of training a pose estimation network with synthetic data. We first provide general information about our model. 
\new{This is followed by a brief description of the training procedure.}
Finally, we detail the student-teacher framework and explain how the teacher is trained. 

{\bf Pose Estimation Network.} For all experiments we use our Tensorflow implementation of the network proposed in \cite{cao2017}, which we will refer to as \emph{OpenPose} network. 
\new{Our training differs only slightly from \cite{cao2017}; details are provided in \supmat}
Because of these differences, we use a self-trained model on real data as the baseline throughout the paper.

{\bf Training with Synthetic Data.} Following the advice of \cite{hinterstoisser2018pre}, we freeze the weights of our feature extractor whenever training on synthetic data. In particular, we freeze the first 4 layers of the OpenPose network.
Additionally, we make sure that each batch is composed of $50\%$ real and $50\%$ synthetic images.
\new{More details on our training procedure with synthetic data are provided in \supmat}

{\bf Grouping.} 
Sampling the most difficult samples with higher probability comes with the risk of oversampling a small amount of data. To avoid such behavior, and ease the task for the teacher, we group the synthetic data into meaningful groups.
We found empirically that the position of the camera and the distance of people in an image contribute to the difficulty of multi-person pose estimation. 
Thus, we use these two image characteristics to group our data.
We assume 10 \emph{groups} to yield a good trade-off between precision and difficulty for the teacher.
For minimal distance grouping\footnote{We sample persons not images. The image is cropped around this person. The minimal distance is defined as smallest distance of this person to any other person.}, denoted as \minDistGroup, we decide for linearly spaced values between $[0,640)$ px. For the camera pitch grouping, denoted as \camPosGroup we space the group boundaries linearly in the interval $[\min(X)+\text{Var}(X),~ \max(X)-\text{Var}(X))$, where $X$ contains all values for the camera pitch in the dataset.

\begin{figure}[t]
    \centering
    \caption{Diagram of the forward pass. The total loss is denoted as $\ell_t$, the loss on real images as $\ell_t^{\mathcal{D}_R}$ and on synthetic images as $\ell_t^{\mathcal{D}_S}$. The reward signal and backward pass are explained in the main text. For clarity only 6 instead of 10 groups are displayed.}
    \includegraphics[width=0.9\textwidth]{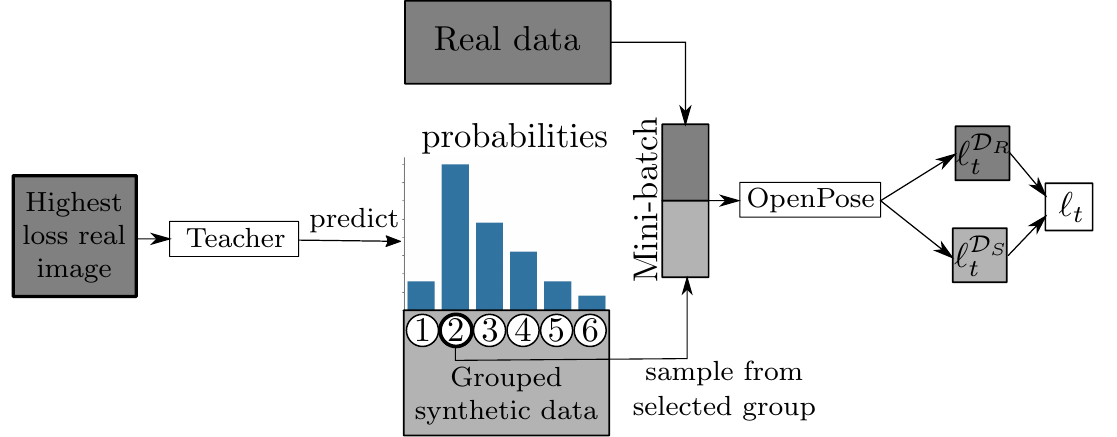}
    \label{fig:pipe_adv}
\end{figure}{}

\subsection{Adversarial Teacher} The teacher network is trained simultaneously with the student network to adapt the sampling strategy dynamically to the current training state of the student.
A schematic of the forward pass can be seen in Fig. \ref{fig:pipe_adv}.
The input to the teacher should represent the training state of the student. 
We choose the real image resulting in highest mean loss per joint within the previous $N$ training steps. 
This provides some information about the type of images that are still difficult for the student.
The output of the teacher network is a probability distribution $\vec{\tilde{P}}$ over a set of groups $\{g_1, \dots, g_i\}$.
This probability distribution is used to sample one of them. For the next $N$ training steps, synthetic training samples are drawn from this group only. 

{\bf Training the Teacher.} The objective of the adversarial teacher should be to maximize the loss, $\ell_t^{\mathcal{D}_S}$. 
Unfortunately, sampling and data augmentation are non-differentiable operations, prohibiting end-to-end training of the teacher network. 
An alternative method to provide a supervision signal is needed.
We draw inspiration from \cite{peng2018jointly} and employ a reward/penalty training scheme.
To determine whether the teacher is rewarded or penalized we monitor $\ell_t^{\mathcal{D}_S}$.
If $\ell_t^{\mathcal{D}_S} \geq \ell_{t-1}^{\mathcal{D}_S}$ the teacher succeeded in finding more difficult examples than before and is rewarded. 
Unfortunately, $\ell_t^{\mathcal{D}_S}$ has a high variance. 
To reduce the variance of the teaching signal we reward if 
\begin{equation}
    	\ell_t^{\mathcal{D}_S} \geq \frac{1}{H} \sum_{h=0}^{H}\ell_{t-1-h}^{\mathcal{D}_S},
	\label{eq:reward_adv}
\end{equation}
where $H$ denotes the number of past loss values to be considered. To avoid favoring images with many people, we use the mean loss per joint on the image.
Eq.~\ref{eq:reward_adv} provides the direction of the gradient descent step but no ground truth is given to compute the gradients.
To efficiently get a reward signal we follow \cite{peng2018jointly} and increase the probability of a group being chosen, if the teacher gets rewarded. Probabilities for other groups are decreased accordingly.
\new{Formally, we update $P_i$ and $P_j$, where $i$ denotes the selected group and $j \neq i$ denotes all other groups, by
\begin{equation}
P_i = \tilde{P}_i + \delta \alpha \tilde{P_i}, \text{ and } P_j= \tilde{P_j} - \delta \frac{\alpha \tilde{P_i}}{|g|-1}.
\label{eq:pseudo_gt_update_i}
\end{equation}}
Here $\tilde{P_i}$ denotes the prediction of the teacher, $P_i$ is the updated pseudo ground truth probability, $0 \le \alpha \leq 1$ controls the size of the update, $|g|$ denotes the number of groups and $\delta$ is a sign indicator $\delta=\{+1, \text{ if Eq. \ref{eq:reward_adv} holds; } -1, \text{ otherwise}\}$.
Finally, we obtain gradients to update the teacher network by computing the KL-divergence loss between $\vec{P}$ and $\vec{\tilde{P}}$. \new{Information on optimization related hyperparameters and the architecture are provided in \supmat}

During training we face a exploration/exploitation trade-off. The group with highest probability, might not be optimal. 
To overcome this problem, we sample a group from a uniform distribution instead of the predicted probabilities with probability $\epsilon=0.1$.

\section{Experiments}

We test our models on the MPII multi-person pose dataset \cite{andriluka14cvpr}. For that purpose we use the toolkit provided by \cite{andriluka14cvpr}.
Following the standard validation procedure, the test metric is only computed for people within close proximity.
\new{We split the real training data, denoted as \dataReal, into a training and a validation set. Our validation set consist of 343 randomly selected groups of people in close proximity. The respective images are not used for training.}
We report mAP  (mean Average Precision), the main metric of the benchmark \cite{andriluka14cvpr} for each model.

\subsection{Which Dataset Generalizes Best?}

We hypothesize that multi-person pose estimation methods are  limited by a lack of training data. To test this hypothesis we train our model on \dataSynth and \dataSynthAndReal.
Interestingly, for \synthOnly, the model resulting from training only on \dataSynth, mAP is very low, suggesting that the model overfits to the synthetic data.
However, when training on \dataSynthAndReal the mAP improves over the mAP of \baseline (Tab.~\ref{tab:baselines}).

    \begin{table*}[t]
		\caption[]{Results on the held out validation set. \baseline is trained only with real data. \synthOnly was trained solely on synthetic data. \realAndSynth is trained on real and synthetic data. \mixed was trained on real and the mixed dataset.}
		\begin{center}
			\footnotesize
			\begin{tabular}{l c c c c c c c c}
				\hline\noalign{\smallskip}
				Model & Head & Shoulder & Elbow & Wrist & Hip & Knee & Ankle & mAP \\
				\noalign{\smallskip}
				\hline
				\noalign{\smallskip}
				\baseline &\textbf{91.3} & 89.1 &79.2 & 70.4 & \textbf{75.9} & 71.5 & 66.7 & 77.7  \\
				\synthOnly 	& 37.9  & 23.5  & 12.7  & 7.3  & 5.6  & 3.4 & 3.2 & 13.4 \\
				\realAndSynth 	& 91.1  & 89.2  & 80.5  & 71.0  & 75.2  & \textbf{73.6} & \textbf{68.1} & \textbf{78.4} \\
				\mixed & \textbf{91.3}  & \textbf{89.5}  &\textbf{ 80.7}  & \textbf{71.7}  & 75.4  & 72.5 & 67.7 & \textbf{78.4} \\ 
				\hline	
			\end{tabular}
			\label{tab:baselines}
		\end{center}
\end{table*}

While synthetic data can improve the accuracy of multi-person pose estimation, the improvements are relatively small given the extensive amount of additional training data. 
Multiple factors might limit the generalization. 
It could well be that the dataset bias between the synthetic and real dataset is just too strong.
Generating the \dataMixed is a straightforward way of generating a dataset with similar dataset bias as the original dataset.
By training on \dataMixed we can quantify the influence of it on the generalization. 
When training with \dataMixed we consider only real humans as samples. The rationale behind that decision is that we primarily want to increase the frequency of occlusion of real humans. However, the network is also trained on all synthetic humans that are within the cropped training image.
As can be seen in Tab.~\ref{tab:baselines}, training on \dataMixedAndReal results in similar accuracy as \realAndSynth.
Therefore, the two methods of generating data are equivalently good.
The dataset bias seems not to be the main limiting factor.

\subsection{Does Stylization Improve Generalization?}

The generalization might be limited by the appearance of synthetic humans.
To overcome the limited generalization and measure how the difference in appearance influences performance we train on \dataStylizedAndReal.
This improves accuracy over \baseline, but leads to a decrease of mAP compared to \mixed (see Tab.~\ref{tab:mixed}). 
This result is surprising, as many factors believed to limit generalization are improved and the data is visually more realistic.
A possible explanation is that the network learns to detect artifacts of the style transfer.
Alternatively, the failure cases of the style transfer method might lead to ``confusion'' of the network.

\begin{table*}[t]
	\begin{small}
		\caption[]{Results for the models \mixed, \stylized, \mixedMasked and \stylizedMasked datasets, where ``+ masks'' denotes masking out loss generated by synthetic humans. All reported results are on the held out validation set.}
		\begin{center}
			\begin{tabular}{l c c c c cc c c}
				\hline\noalign{\smallskip}
				Model & Head & Shoulder & Elbow & Wrist & Hip & Knee  & Ankle & mAP\\
				\noalign{\smallskip}
				\hline
				\noalign{\smallskip}	
				\mixed &  91.3  & 89.5  & 80.7  & 71.7  & 75.4  & \textbf{72.5} & 67.7 & 78.4 \\
		    	\mixedMasked & \new{\textbf{92.3}}  & \new{\textbf{90.9}}  & \new{80.5}  & \new{\textbf{72.2}}  & \new{76.0}  & \new{71.7} & \new{68.3} & \new{78.9} \\ 
				\stylized & \new{91.8}  & \new{89.8}  & \new{80.4}  & \new{70.9}  & \new{75.5}  & \new{71.6} & \new{67.9} & \new{78.3} \\ 
    				 \stylizedMasked & \new{91.6}  & \new{90.6}  & \new{\textbf{80.8}}  & \new{71.8}  & \new{\textbf{77.7}}  & \new{72.2} & \new{\textbf{68.8}} & \new{\textbf{79.1}} \\ 
             \hline	
			\end{tabular}
			\label{tab:mixed}
		\end{center}
	\end{small}
\end{table*}	    

In an ablation study, we test whether training on synthetic humans actually improves the mAP or if the improvement when training with \dataMixed or \dataStyle is mostly caused by additional occlusion. 
For that purpose, we mask out all the loss that is generated by synthetic humans. 
\new{Comparing \mixedMasked with \mixed, it can be seen that masking out the loss generated by synthetic humans increases the accuracy.
Therefore, the domain gap between synthetic and real humans limits the generalization, and the improvement of \mixed over \baseline is due to more occlusion.}
\new{Stylization in combination with masks leads to the best model (see Fig.~\ref{fig:teaser} and \supmat for qualitative results), suggesting that a smaller gap between synthetic occluder and real parts of the image improves generalization.}

\subsection{Does Informed Sampling Improve Results?}
    
Finally, we test whether the teacher network can help to use the synthetic data more effectively.
For that purpose we use the adversarial teacher with \dataSynthAndReal.
The results can be seen in Tab.~\ref{tab:pol_adv_best}. 
Grouping according to the camera \new{pitch} leads to an additional improvement of $0.5$ mAP in comparison to \realAndSynth.
See Fig.~\ref{fig:teaser} and \supmat for qualitative results. 
\new{For usage of the teacher in combination with the mixed and stylized datasets we do not observe further improvements. Since we only consider real humans as samples some groups are very small. We assume that oversampling of these groups inhibits improvements.}

\begin{table*}[t]
	\begin{small}
		\caption[]{
		            Comparison of the \baseline and \realAndSynth baselines (copied from Tab.~\ref{tab:baselines}) to the student-teacher model using different groupings. 
		            Results on the validation set.
		}
		\begin{center}
			\begin{tabular}{l c c c c c c c c c}

\hline	\noalign{\smallskip}

				Model & Grouping & Head & Shoulder & Elbow & Wrist & Hip & Knee & Ankle & mAP \\
				\noalign{\smallskip}
				\hline
				\noalign{\smallskip}
				\baseline & & 91.3 & 89.1 &79.2 & 70.4 & 75.9 & 71.5 & 66.7 & 77.7  \\
				\realAndSynth & & 91.1  & 89.2  & 80.5  & 71.0  & 75.2  & 73.6 & \textbf{68.1} & 78.4 \\		
				\advTeacherTd & \camPosGroup & \textbf{91.7}  & 90.0 & \textbf{80.9}  & 71.2  & \textbf{77.1}  &\textbf{73.6} & 67.7 & \textbf{78.9} \\
				\advTeacherTd & \minDistGroup & 91.5  & \textbf{90.4}  &80.5  & \textbf{72.2}  & 75.8  & 73.1 & 67.6 & 78.7 \\
				\hline	
			\end{tabular}
			\label{tab:pol_adv_best}			
		\end{center}
	\end{small}
\end{table*}

\begin{figure}[t]
    \centering
    \caption{Detection performance for varying ratio of visible joints (mAP). The validation data is grouped into 5 linearly spaced groups in range $[0,1]$. The groups contain $14$, $111$, $289$, $286$, $180$ persons respectively. The teacher methods are hard to distinguish because of similar values.}
    \includegraphics[width=0.65\textwidth]{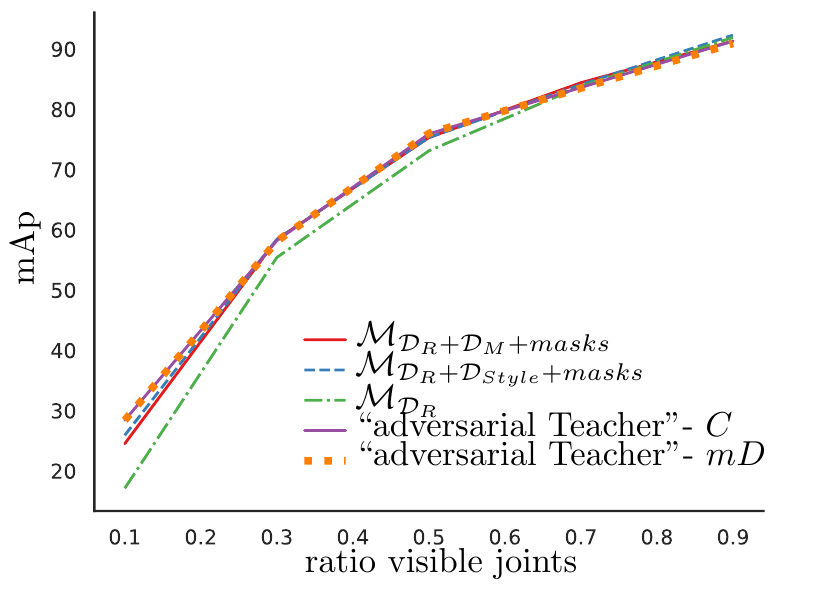}
    \label{fig:occlevels}
\end{figure}

As can be seen in Fig.~\ref{fig:occlevels}, improvements for highly occluded people are strongest for models trained with the teacher. \new{Clear differences to \mixedMasked and \stylizedMasked can only be seen for the highest occlusion level. For all but the lowest occlusion levels \baseline is outperformed by all other models}. Thus, our method\new{s} of training improve accuracy for difficult high occlusion cases.

{}

\new{{\bf Sampling Probabilities. }The teacher often assigns high probability to few groups early in training. In most cases the teacher converges to a uniform sampling strategy over the groups, as training progresses. This is not equivalent to random sampling, as samples are not uniformly distributed across groups.
As a result the training data follows a uniform distribution for the respective image characteristic. 
A uniform distribution over an image characteristic like camera pitch is more extreme than the distribution in the real training data.
Training on more extreme samples seems to improve the generalization. We find that improvements are largest for the camera pitch grouping. This might be due to the bias to small camera pitch in the real training data.}
\section{Conclusion}

In summary, we created multiple synthetic datasets and analyze their effectiveness. 
We show that training with synthetic data improves multi-person pose estimation methods. 
We find that both our methods of generating synthetic datasets perform on par.
Surprisingly, our approach for improving visual appearance of synthetic humans decreased the accuracy.
More elaborate domain adaptation methods might provide better results. For example, GAN-based approaches ensure that no obvious artifacts can be used to distinguish real from synthetic. 
We find that improvements of the \mixed and \stylized can be explained by more occlusion, as mAP increases when masking out the loss of synthetic humans. \new{Here the stylization leads to further improvements, suggesting that visual quality of occluding objects is important.}
Finally, we find that training on the most difficult synthetic samples at each point of training improves the results. This suggest that, for large synthetic datasets, random sampling is not optimal and better strategies exist.
More research in this direction is necessary to draw final conclusions.
\new{\stylizedMasked outperforms the \advTeacherTd \camPosGroup model. 
We assume that the potential of the teacher is limited by the small amount and the quality of human textures. Given better textures we expect this approach to outperform \stylizedMasked.}

{\bf Limitations. }The teacher network is limited in multiple ways.
First, the current implementation requires grouping of data. 
The size and spacing of groups might have a large influence on the training success and applicability of teacher networks. 
Furthermore, the grouping is based on one feature only.
This is suboptimal, since difficulty of an image is determined by multiple factors. We plan to extend the teachers to handle multiple characteristics at once.
A more elaborate formulation that does not require grouping might be superior.
\new{Last, the teacher can be applied to other tasks and networks, here we evaluate it only for multi-person pose estimation with the OpenPose network.}

The style transfer occasionally produces artifacts in the stylized image.
Especially the skin color of the synthetic humans might be unnatural.
In rare cases, large parts of the background are included in the human mask. 
These failures in segmentation can lead to ghost-like synthetic humans (see \supmat).

\vspace{0.5em}
\noindent
\new{
{\bf Acknowledgement.}
S.~Tang acknowledges funding by Deutsche Forschungsgemeinschaft (DFG, German Research Foundation) Projektnummer 276693517 SFB 1233. \\
{\bf Disclosure.}
MJB has received research gift funds from Intel, Nvidia, Adobe, Facebook, and Amazon. While MJB is a part-time employee of Amazon, his research was performed solely at, and funded solely by, MPI. MJB has financial interests in Amazon and Meshcapade GmbH.
}

\newpage
\bibliographystyle{splncs04}
\bibliography{egbib}
\clearpage



\section*{Supplementary material}

# Learning to Train with Synthetic Humans
# **Supplementary Material**

David T. Hoffmann[1], Dimitrios Tzionas[1], Michael J. Black[1], and Siyu Tang[1,2,3]

[1] Max Planck Institute for Intelligent Systems, Germany
[2] University of Tübingen, Germany
[3] ETH Zürich
`dhoffmann, dtzionas, black, stang@tuebingen.mpg.de`

## 1 Data Generation

**MoCap Data.** To pose the body we use $1,515$ MoCap sequences from the CMU dataset [1], 62 from HumanEva [14], 41 from the PosePrior dataset [3] and 157 unpublished sequences recorded with our own motion capture system. To reduce the similarity of poses we subsample every 10th frame, resulting in 12 fps for most of the datasets. We end up with $253,762$ individual poses.

**Hand Poses.** Without variations of hand poses, a keypoint detector trained on the synthetic data might not generalize to other hand poses. To avoid this, we use SMPL+H [13] and pose the hands and fingers. However, conventional MoCap systems do not record the pose of fingers [1, 3, 14]. To obtain realistic poses for hands and fingers we use the "embodied hands" dataset [13].

**Shape.** Besides body pose, humans differ in their body shape. To express these differences in our synthetic datasets we extract shape parameters $\beta$ from standard MoCap datasets using MoSh [9].

**Textures.** Textures have a large influence on the perceived realism of synthetic data. For the synthetic humans we use the textures published with the SURREAL dataset [18]. The dataset provides 772 textures of people in casual clothing and 157 in minimal clothing. The former were collected by the authors of [18]. The latter were acquired from the CAESAR dataset [11]. We use 80% of these textures and keep the remaining 20% for validation and test data, even though we do not use any synthetic validation or test data for this particular work. The reason for this decision is to keep the data splits for training, test and validation consistent across projects.

**Noise and Lighting.** Real images are subject to multiple sources of noise, for example motion blur or defocus. Furthermore, they are taken under varying lighting conditions. We model noise with Gaussian blur. We blur x and y direction independently with a probability of 0.5, each. The size of the Gaussian kernel in pixels is given by the absolute value sampled from a standard normal distribution. We vary lighting conditions similarly as [18, 10] using spherical harmonics [6] with randomly drawn coefficients.

**Sampling Motions.** Sampling MoCap sequences randomly does not guarantee that every MoCap sequence is used. To ensure that, we select the sequence

for the first synthetic human deterministically. MoCap sequences for all other synthetic humans are sampled randomly. The probability of sampling a MoCap sequence $S_j$ is given by $p_j = \frac{|S_j|}{\sum_{i=1}^{|S|} |S_i|}$. Here $|S_i|$ denotes the number of frames of sequence $i$ and $|S|$ the number of sequences. Since the chosen sequences might have different length, we further randomly sample for each one the frame ID for the starting frame, to encourage the use of the whole sequences.

**Changing Rendering Parameters.** To increase variance in the dataset multiple parameters are changed. For each MoCap sequence we change the camera position, number of humans and the global rotation of each synthetic human. In contrast, background image and lighting as well as position, pose, shape and texture of synthetic humans are changed for each frame.

**Scene Generation.** After sampling all these parameters, each virtual human is placed on an invisible ground plane in the field of view of the camera. Maximal distance to the camera are 12 m.

**Collision Detection.** The meshes of synthetic humans in the 3D scene might intersect, resulting in physically impossible configurations. A quick and easy way of detecting collisions are axis aligned bounding boxes. However, their usage results in a large number of false positives and limits the distances between virtual humans. For multi-person pose estimation, however, small distances between humans are frequent and should be represented in the training data. We draw inspiration from [17] and use bounding volume hierarchies (BVH) [16] instead. This is an efficient method to check for collisions on triangle level. Thus, it allows for smallest possible distances between the meshes.

Whenever two synthetic humans collide, we search for a new, valid position for one of them before rendering the image. Frames with mesh self-collisions are not rejected, as they are very frequent for highly articulated poses, which might be important for a pose-estimation dataset.

## 2   Datasets

Tab. A.1 shows a comparison of our datasets to related other datasets. JTA and SURREAL are much larger, however SURREAL only considers a single person in indoor environments and JTA only urban scenes in surveillance scenarios. Our datasets have a much higher variety in terms of scenes.

Example images for $\mathcal{D}_S$ can be seen in Fig. A.1

### 2.1   $\mathcal{D}_M$ and $\mathcal{D}_{Style}$ - Qualitative Results

Fig. A.2 and A.3 show example images of $\mathcal{D}_M$ and the corresponding images from $\mathcal{D}_{Style}$. In particular Fig. A.2 (A) shows that the style transfer method generates realistic variations of textures and changes their color. In Fig. A.2 (B, C) and Fig. A.3 (A, B, C) it can be seen that the method adapts the lighting of textures to fit better into the scene. It greatly improves blending-in of synthetic humans. Fig A.2 (D) and Fig A.3 (D) show failure cases. Here, larger parts of the background were included in the human mask of mask-RCNN [7, 2]. As a

**Fig. A.1.** Example images from $\mathcal{D}_S$. The dataset contains images with interesting poses, challenging backgrounds, variance in camera position and heavy occlusion.

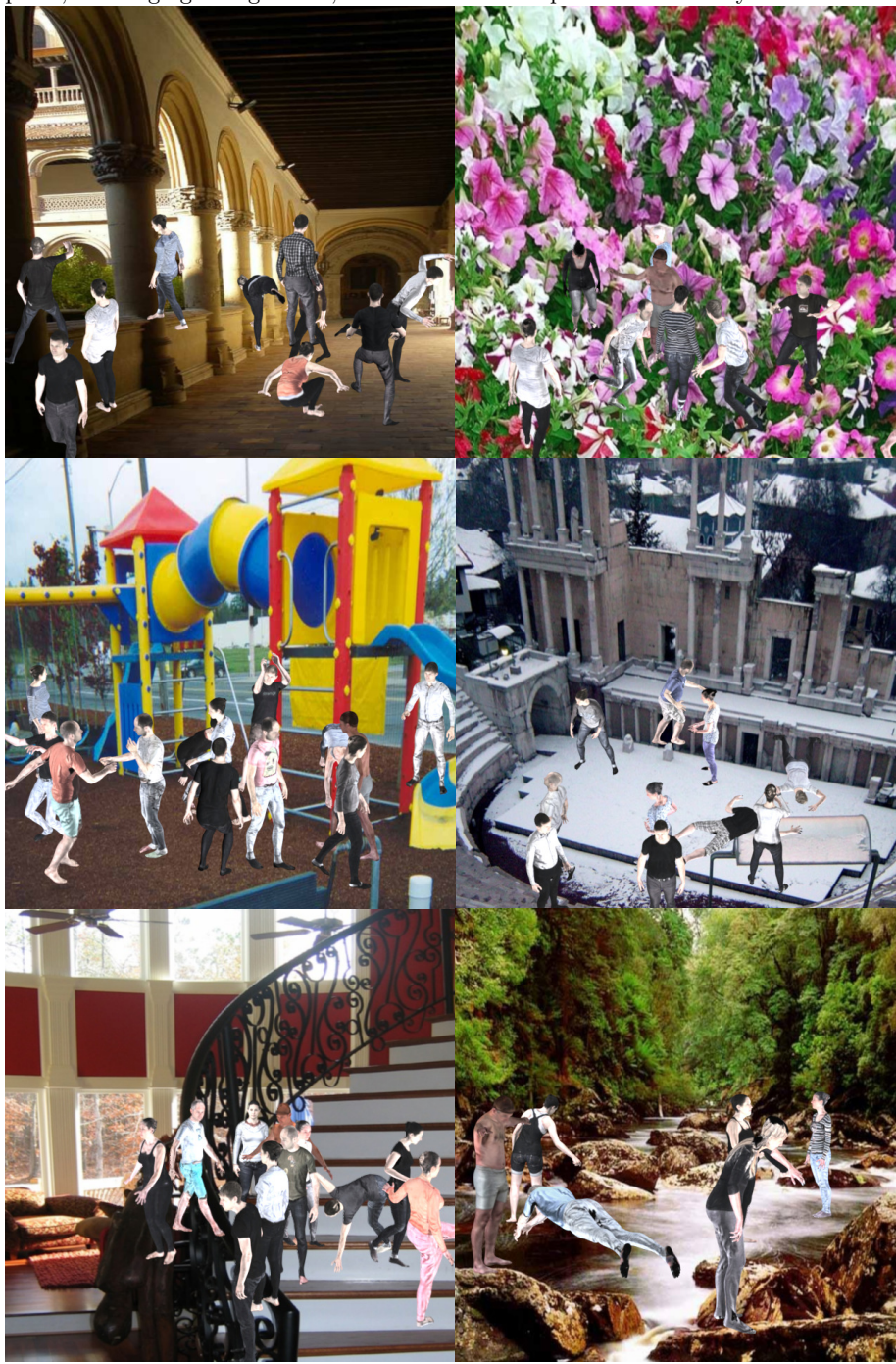

**Table A.1.** Number of frames and poses for our datasets and for the most related synthetic datasets. All datasets are multi-person datasets, except for SURREAL. For $\mathcal{D}_M$ and $\mathcal{D}_{Style}$ "#Synthetic Humans" corresponds to the additional number of synthetic humans. For $\mathcal{D}_S$, SURREAL, JTA and SURREAL-style multi-person "#Synthetic Humans" corresponds to the number of annotated poses.

| Dataset | | #frames | #Synthetic Humans |
|---------|------|---------|-------------------|
| $\mathcal{D}_S$ | ours | 70,379 | 580,693 |
| $\mathcal{D}_M$ | ours | 74,628 | 279,605 |
| $\mathcal{D}_{Style}$ | ours | 74,352 | 279,278 |
| SURREAL | [18] | 6,536,752 | 6,536,752 |
| JTA | [5] | 460,800 | 10,000,000 |
| SURREAL-style multi-person | [12] | 40,000 | 186,000 |

result, synthetic humans are partially stylized in the style of the background. This leads to the observable ghost-effect.

## 3    Training

**Pose Estimation Network.** Unfortunately not all hyper-parameters used for training on the MPII pose estimation dataset were provided for the OpenPose network. Our hyper-parameter search results in a model that approaches the original performance. Differences in performance may be due to a different choice of optimizer. We use the more common Adam algorithm [8], whereas Cao et al. [4] rely on SGD. Our hyperparameter search lead to a learning rate of $lr = 0.0001$, $\beta_1 = 0.8$ and $\beta_2 = 0.999$. Furthermore, we found that a learning rate decay improves results. We decay the learning rate every 20.000 steps with a decay rate of 0.66. We use a batch size of 32.

Most of our models are initialized with weights pretrained on real data. The only 2 models that are not pretrained on real pose estimation data are $\mathcal{M}_{\mathcal{D}_R}$ and $\mathcal{M}_{\mathcal{D}_S}$. For these models we follow the same procedure as proposed by Cao et al. [4] and initialize them with the first 10 layers of VGG-19. The remaining weights are randomly initialized.

**Training with Synthetic Data.** All models trained with synthetic data are initialized with pretrained weights. These weights are obtained by training OpenPose for 70.000 steps on real data. Whenever we start from these pretrained weights, we use an initial learning rate of $lr = 0.00005$.

### 3.1    Data Augmentation

To increase the number of training samples we apply standard data augmentation techniques. We implement the same data augmentation pipeline as Cao et al. [4], however the hyperparameters might be different. We scale the image in the range $[0.4, 1.6]$, rotate it by a uniformly sampled value in the interval $[-45°, 45°]$.

**Fig. A.2.** Example images from $\mathcal{D}_M$ and the respective image from $\mathcal{D}_{Style}$. Last row shows a failure case with ghost-like appearance of synthetic humans.

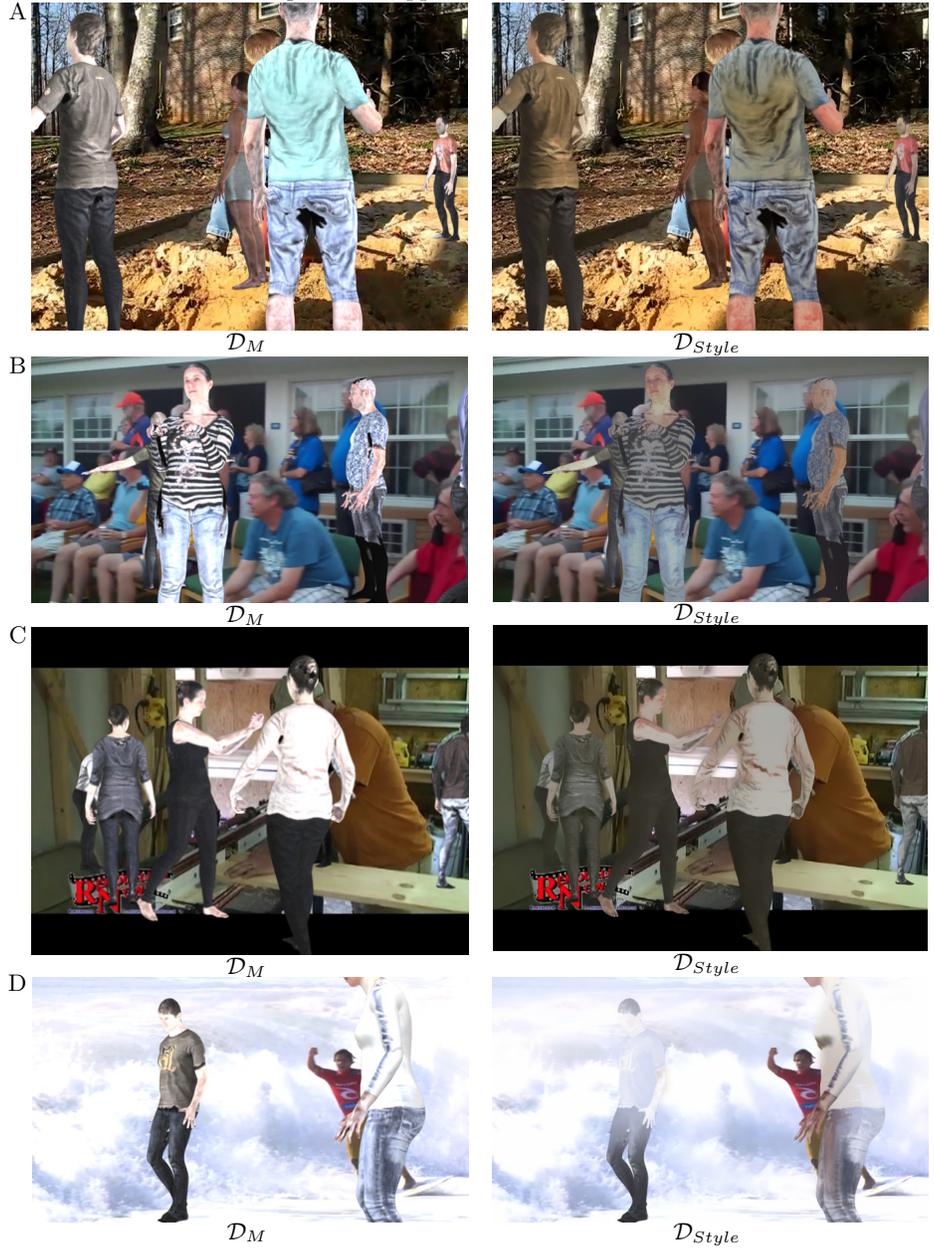

**Fig. A.3.** Example images from $\mathcal{D}_M$ and the respective image from $\mathcal{D}_{Style}$. Last row shows a failure case with ghost-like appearance of synthetic humans.

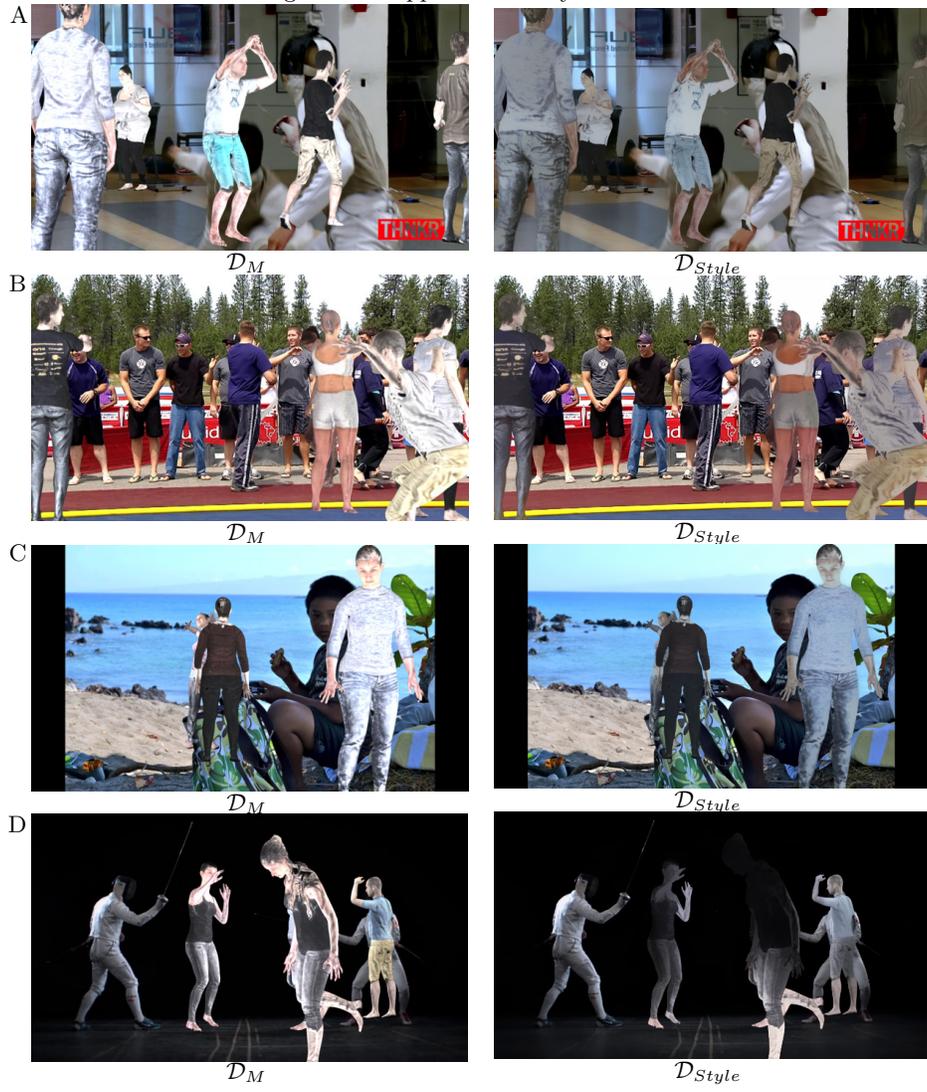

Furthermore, each image is cropped around a target person. Size of the crop is $368 \times 368$ px. We add noise to the center of the crop. It is uniformly sampled from $[-50, 50]$ px. Finally, with probability of 0.5 the image is flipped horizontally.

### 3.2   The Teacher Network

To optimize the teacher we use a constant learning rate of 0.00005. Similar to optimization of the OpenPose network we use the Adam algorithm with $\beta_1 = 0.8$ and $\beta_2 = 0.999$.

**Architecture.** The teacher network consists of two parts. The first part are the first 10 layers of VGG-19 [15] and is identical to the OpenPose feature extractor. Similar to the OpenPose feature extractor the weights of VGG-19 are used to initialize it. The second part of the teacher network differs from the OpenPose network. It consists of $3 \times 3 \times 256$ and $3 \times 3 \times 128$ convolutional layers, followed by two $3 \times 3 \times 64$ layers. After every two layers we add a max pooling layer. The last two layers are fully-connected with 512 units each.

## 4   Multi-Person Pose Estimation - Qualitative Results

Qualitative results for $\mathcal{M}_{\mathcal{D}_R}$, "adversarial Teacher" $C$ and $\mathcal{M}_{\mathcal{D}_R + \mathcal{D}_{Style} + \mathrm{masks}}$ can be seen in Fig. A.4, Fig. A.5 and Fig. A.6. While a clear improvement can be seen by "adversarial Teacher" $C$ and $\mathcal{M}_{\mathcal{D}_R + \mathcal{D}_{Style} + \mathrm{masks}}$ over $\mathcal{M}_{\mathcal{D}_R}$, differences between "adversarial Teacher" $C$ and $\mathcal{M}_{\mathcal{D}_R + \mathcal{D}_{Style} + \mathrm{masks}}$ are more subtle. We found that their predictions are qualitatively on par for most images.

As can be seen in Fig. A.4 (A, B) and Fig. A.6 (A) grouping of joints fails frequently for $\mathcal{M}_{\mathcal{D}_R}$. In contrast, erroneous grouping can be less frequently observed for our models. Thus, the models trained on our datasets improve in their ability to group joints. For very crowded scenes as shown in Fig. A.4 (A) and Fig. A.6 (A) the "adversarial Teacher" $C$ seems to outperform $\mathcal{M}_{\mathcal{D}_R + \mathcal{D}_{Style} + \mathrm{masks}}$. Thus, adversarial training on the more crowded purely synthetic dataset is beneficial for real crowded scenes.

Besides grouping, training on our datasets seems to improve the detection of occluded people and occluded keypoints. Examples for this can be seen in Fig. A.4 (A, B, C), Fig. A.5 (B, C) and Fig. A.6 (A). Of particular interest is Fig. A.5 (A). Here it can be seen that the training on purely synthetic data with adversarial teacher leads to better detection of people under challenging imaging conditions and improves the detection and grouping of joints in front of highly cluttered backgrounds.

Last, Fig. A.6 (A, B) suggest that "adversarial Teacher" $C$ improves the prediction for uncommon camera positions. Thus, training on the most challenging camera positions improves the predictions for such images in comparison to $\mathcal{M}_{\mathcal{D}_R}$ and $\mathcal{M}_{\mathcal{D}_R + \mathcal{D}_{Style} + \mathrm{masks}}$.

**Fig. A.4.** Example images with detected poses for $\mathcal{M}_{\mathcal{D}_R}$, the model trained with adversarial teacher using the camera pitch grouping ("adversarial Teacher" $C$) and our best model $\mathcal{M}_{\mathcal{D}_R + \mathcal{D}_{Style} + \text{masks}}$.

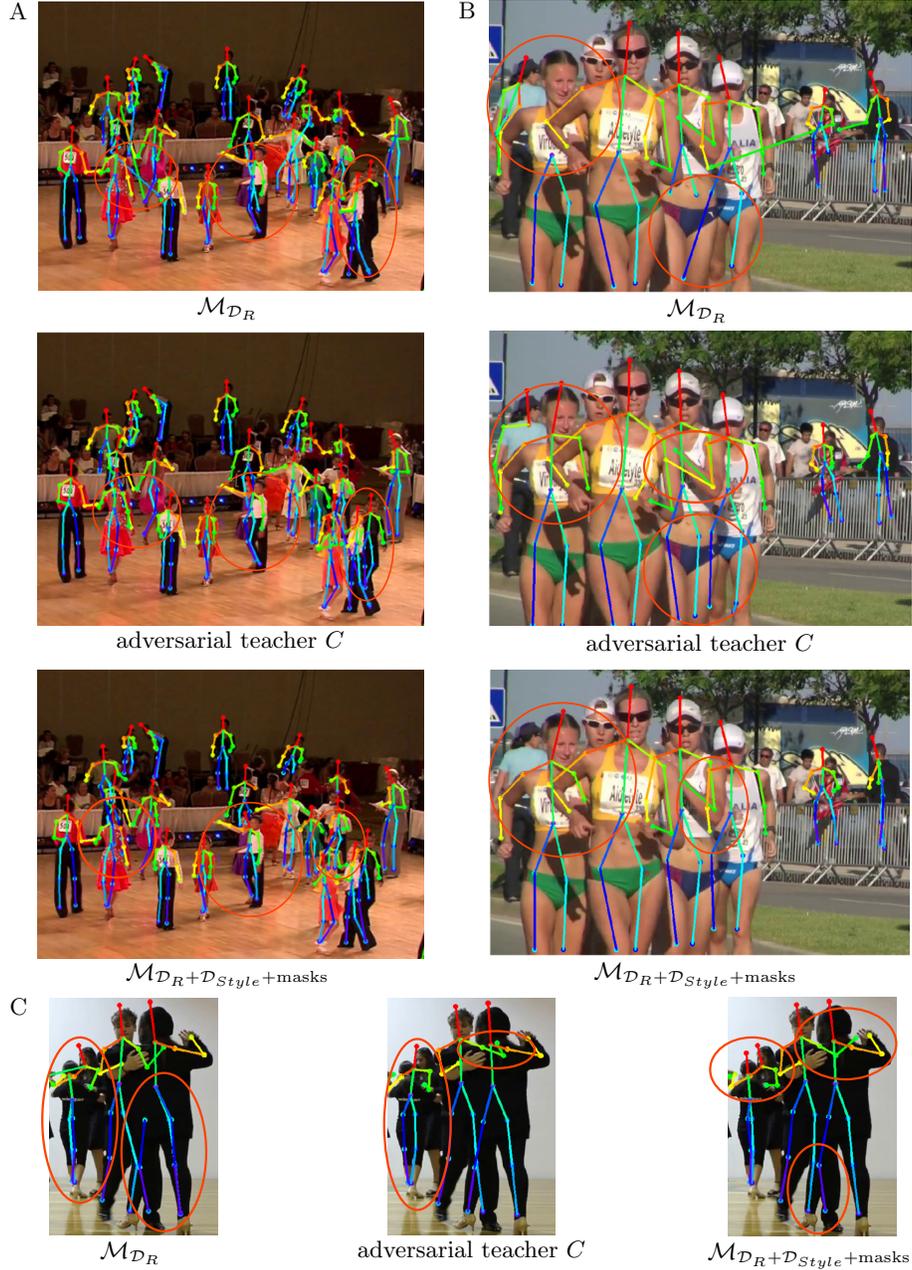

**Fig. A.5.** Example images with detected poses for $\mathcal{M}_{\mathcal{D}_R}$, the model trained with adversarial teacher using the camera pitch grouping ("adversarial Teacher" $C$) and our best model $\mathcal{M}_{\mathcal{D}_R + \mathcal{D}_{Style} + \text{masks}}$.

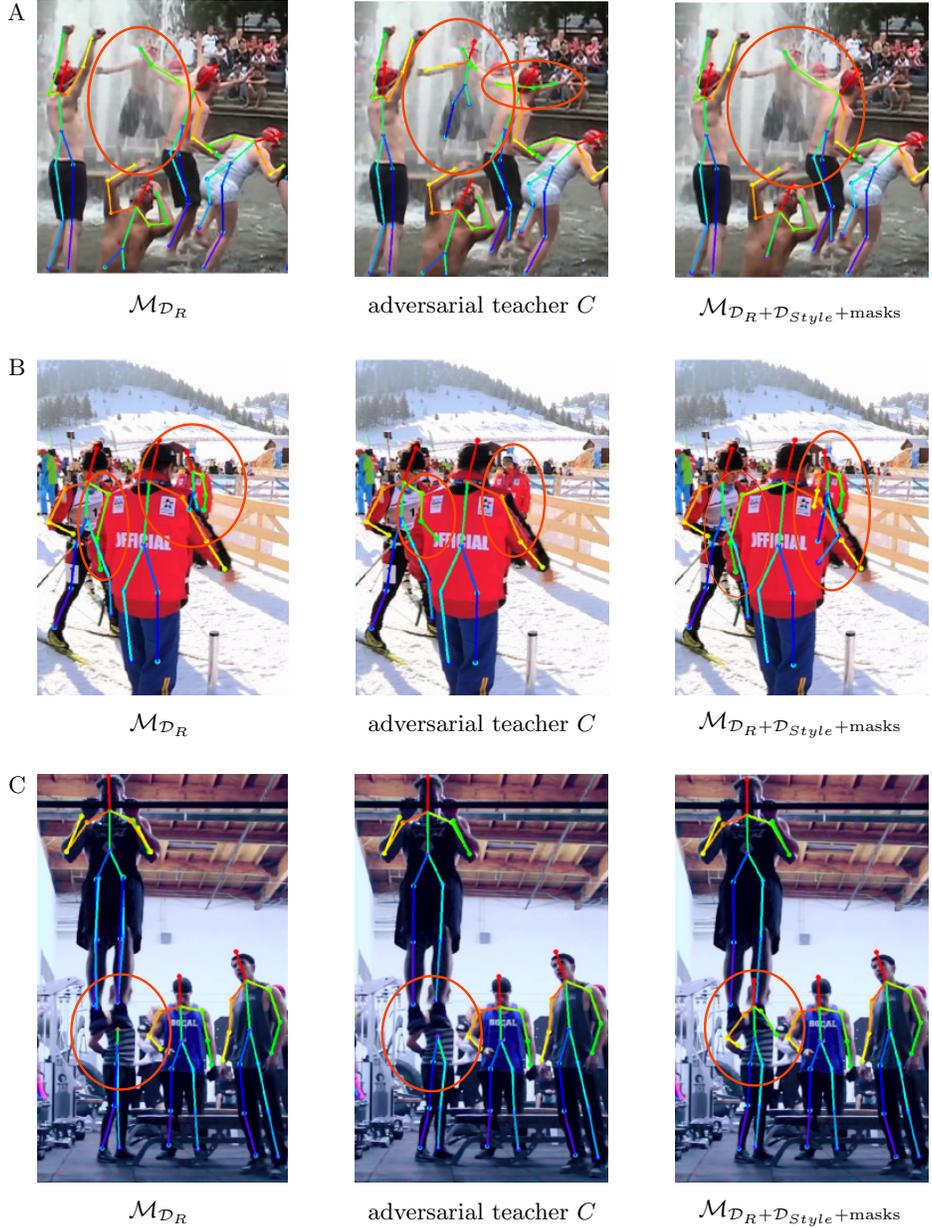

**Fig. A.6.** Example images with detected poses for $\mathcal{M}_{\mathcal{D}_R}$, the model trained with adversarial teacher using the camera pitch grouping ("adversarial Teacher" $C$) and our best model $\mathcal{M}_{\mathcal{D}_R+\mathcal{D}_{Style}+\text{masks}}$.

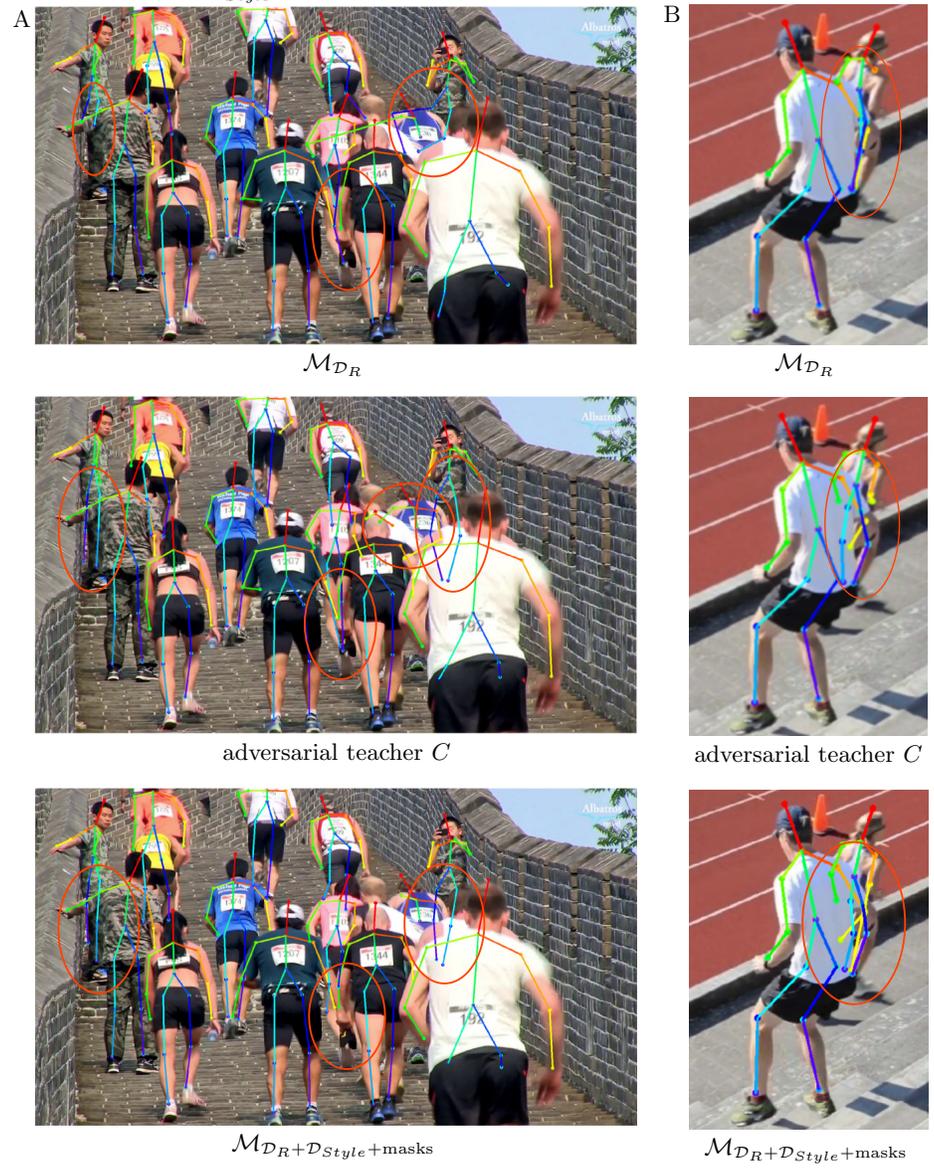

A

$\mathcal{M}_{\mathcal{D}_R}$

B

$\mathcal{M}_{\mathcal{D}_R}$

adversarial teacher $C$

adversarial teacher $C$

$\mathcal{M}_{\mathcal{D}_R+\mathcal{D}_{Style}+\text{masks}}$

$\mathcal{M}_{\mathcal{D}_R+\mathcal{D}_{Style}+\text{masks}}$


\end{document}